\documentclass[12pt]{article}

\usepackage{rsipacks} 

\usepackage{float}  
\usepackage{subcaption}

\usepackage[font=it]{caption}
\theoremstyle{definition}

\usepackage{graphicx}
\theoremstyle{remark}

\sectionfont{\large \centering}
\subsectionfont{\normalsize \centering}
\crefname{claim}{claim}{claims}
\Crefname{claim}{Claim}{Claims}
\crefname{app-corollary}{corollary}{corollaries}
\Crefname{app-corollary}{Corollary}{Corollaries}
\crefname{app-definition}{definition}{definitions}
\Crefname{app-definition}{Definition}{Definitions}
\crefname{figure}{figure}{figures}
\Crefname{figure}{Figure}{Figures}
\crefname{lemma}{lemma}{lemmata}
\Crefname{lemma}{Lemma}{Lemmata}
\crefname{app-lemma}{lemma}{lemmata}
\Crefname{app-lemma}{Lemma}{Lemmata}
\crefname{app-proposition}{proposition}{proposition}
\Crefname{app-proposition}{Proposition}{Proposition}
\crefname{app-theorem}{theorem}{theorems}
\Crefname{app-theorem}{Theorem}{Theorems}

\begin{document}
\title{Deepfake Geography: Detecting AI-Generated Satellite Images}
\author{Mansur Yerzhanuly}
\date{}
\maketitle

\begin{abstract}
\noindent The rapid advancement of generative models such as StyleGAN2 and Stable Diffusion poses a growing threat to the authenticity of satellite imagery, which is increasingly vital for reliable analysis and decision-making across scientific and security domains. While deepfake detection has been extensively studied in facial contexts, satellite imagery presents distinct challenges, including terrain-level inconsistencies and structural artifacts. In this study, we conduct a comprehensive comparison between Convolutional Neural Networks (CNNs) and Vision Transformers (ViTs) for detecting AI-generated satellite images. Using a curated dataset of over 130,000 labeled RGB images from the DM-AER and FSI datasets, we show that ViTs significantly outperform CNNs in both accuracy ($95.11\%$ vs. $87.02\%$) and overall robustness, owing to their ability to model long-range dependencies and global semantic structures. We further enhance model transparency using architecture-specific interpretability methods, including Grad-CAM for CNNs and Chefer’s attention attribution for ViTs, revealing distinct detection behaviors and validating model trustworthiness. Our results highlight the ViT’s superior performance in detecting structural inconsistencies and repetitive textural patterns characteristic of synthetic imagery. Future work will extend this research to multispectral and SAR modalities and integrate frequency-domain analysis to further strengthen detection capabilities and safeguard satellite imagery integrity in high-stakes applications.

\end{abstract}

\section{Introduction}
\noindent Satellite imagery plays a crucial role in how governments and media outlets observe and interpret the world \cite{zhu2017deeplearning}. From tracking deforestation to monitoring urban development and conflict zones, remote sensing data is widely trusted as an objective source of truth. However, recent developments in generative AI have introduced a significant threat to the authenticity of this data. With tools such as StyleGAN2 and Stable Diffusion, it is now possible to generate synthetic satellite images that are visually convincing yet entirely fictional. These fake images have the potential to distort evidence, fabricate geographical events, or spread misinformation at a large scale \cite{10282927}\cite{zhao2021deepfake}.\\\\
While deepfake detection in facial imagery has received substantial research attention, most existing methods rely on human-specific cues such as blinking patterns, skin textures, or facial dynamics \cite{8630787}\cite{9141516}. These cues are absent in satellite data, making direct adaptations of facial detection models ineffective. The problem of detecting deepfakes in satellite imagery is fundamentally different and more complex. Instead of localized anomalies, satellite forgeries often contain subtle, large-scale inconsistencies such as repeated terrain textures, unnatural transitions, and mismatched shadows, artifacts that are harder to detect and less explored in the current literature \cite{10282927}\cite{zhang2019detecting}.
\begin{figure}[H]
    \centering
    \includegraphics[width=0.8\linewidth]{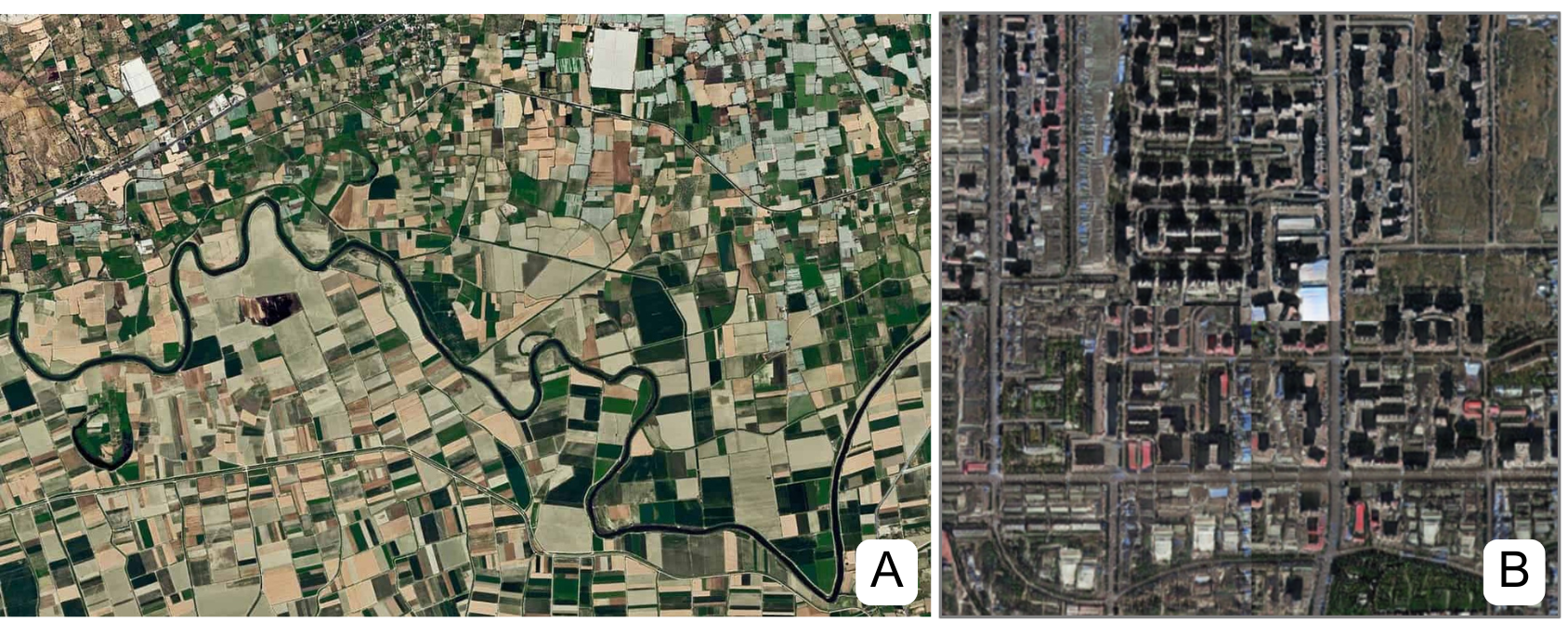}
    \caption{Comparison between a real rural satellite image (A, source: IndiaAI, 2023 \cite{indiaai2023deepfakemaps}) and a synthetic urban image (B, source: GeekWire, 2021 \cite{yonck2021deepfakegeography}), illustrating natural terrain continuity (A) versus repeated textures and abrupt transitions typical of AI-generated imagery (B).}
    \label{fig:real_vs_fake}
\end{figure}
\noindent
This research investigates how deep learning can be used to detect AI-generated satellite images. We focus on two leading architectures for image classification: Convolutional Neural Networks (CNNs), which have shown success in visual forensics, and Vision Transformers (ViTs), which are well-suited for modeling long-range dependencies and holistic image structure. To support our experiments, we curated a combined dataset of approximately 134,000 satellite images from the DM-AER \cite{deepmedia2022dmaer} and FSI \cite{zhao2021deepfake} datasets, encompassing both real and AI-generated samples. This provided a diverse and representative foundation for training and evaluating our models.\\\\
In addition to performance metrics such as accuracy and F1-score, we apply interpretability tools, including Grad-CAM and transformer attribution maps, to analyze where and how models detect visual and structural anomalies. Our work contributes to a growing field of geographic image forensics by (1) conducting a systematic comparison between CNNs and ViTs, (2) identifying visual and frequency-domain artifacts unique to generative satellite imagery, and (3) proposing a set of explainable benchmarks for future research. The methods and insights developed in this project have practical applications in journalism, environmental science, and defense, where verifying the authenticity of satellite images is increasingly important.\\\\
To preview our findings, we show that the Vision Transformer (ViT-B/16) outperformed the CNN (ResNet-50), achieving a $95.11\%$ test accuracy and demonstrating greater robustness to distributional shifts. While CNNs effectively captured localized anomalies, ViTs excelled at detecting broader structural inconsistencies such as unnatural terrain transitions and semantic dissonance. These results underscore the importance of modeling long-range dependencies for reliable deepfake detection in satellite imagery.\\\\
\section{Related Works} \label{sec:setting}
\noindent \textbf{2.1. Deepfake Detection: From Facial to Satellite Imagery}\\\\
Most deepfake detection research has focused on facial imagery, where models leverage physiological signals such as eye blinking \cite{8630787} or subtle cardiac rhythms \cite{9141516}. These approaches, while effective in portrait-based contexts, do not translate to satellite imagery for two key reasons: (1) the absence of physiological features like facial landmarks or heartbeats, and (2) the presence of fundamentally different artifact patterns, such as landscape-level inconsistencies.\\\\
Facial detectors typically identify localized anomalies; for instance, lip or eye swaps, whereas satellite forgeries often exhibit large-scale irregularities, including unnatural texture repetitions or shadow mismatches \cite{10282927}.
\\\\
Recent surveys \cite{mirsky2021deepfakes}\cite{tariq2018detecting} highlight the promise of blind detection methods, techniques that operate without domain-specific priors. These models offer greater flexibility across different image types. However, because our models are trained exclusively on satellite data, they are not truly blind or domain-agnostic. Instead, our work remains domain-specific, focusing on forensic patterns unique to synthetic satellite imagery.
\\\\
\textbf{2.2. Satellite-Specific Deepfake Detection}\\\\
Deepfake detection specific to satellite images is still an emerging area of research and lacks the maturity and volume of work seen in facial forgery studies, which often build on years of benchmark datasets and well-defined physiological features.\\\\
A recent comprehensive survey by Ding et al. \cite{ding2024forensic} further underscores this gap, emphasizing that satellite image forgery forensics remains in its early stages and faces challenges such as the scarcity of standardized datasets and robust evaluation benchmarks.
\\\\
A notable contribution in this space is the model by \c{C}ift\c{c}i and Demir \cite{10282927}, which uses a multi-attention super-resolution framework and achieves $99.5\%$ accuracy on the DM-AER and FSI datasets. Their patch-based strategy is effective at detecting global inconsistencies. However, their approach has two limitations. First, it does not compare traditional CNNs and Vision Transformers (ViTs), leaving open questions about architectural performance. Second, while attention maps are used for explanation, they remain qualitative and fail to precisely localize artifacts such as shadow mismatches or terrain duplications.
\\\\
While Zhao H., Zhou S., et al. \cite{zhao2021multiattentionaldeepfakedetection} showed that attention mechanisms improve deepfake detection in facial images, adapting these methods to satellite-specific cues, such as terrain continuity, road logic, and spatial coherence, has not been thoroughly explored. This is a gap our study seeks to address.
\\\\
\textbf{2.3. Vision Transformers for Geographic Forensics}\\\\
Vision Transformers (ViTs), introduced by Dosovitskiy et al. \cite{dosovitskiy2021imageworth16x16words}, are well-suited to satellite deepfake detection due to their ability to model relationships across distant regions of an image. This capacity for capturing global spatial structure enables ViTs to recognize large-scale anomalies that CNNs may miss.\\\\
In natural image forensics, ViTs have outperformed CNNs in tasks like tampering localization \cite{wang2022m2tr}. We hypothesize that ViTs will similarly excel in satellite imagery, particularly in identifying: (1) structural inconsistencies, such as illogical river bends or disconnected road networks, and (2) global texture repetitions, like cloned forest patches or duplicated terrain.
\\\\
Traditional remote sensing models often focus on tasks like object detection or land classification \cite{zhu2017deeplearning}\cite{LI2020296} but rarely consider the forensic problem of detecting synthetically generated imagery. Our work aims to fill this gap.\\\\
\textbf{2.4. Artifacts in Generative Satellite Imagery}\\\\
Generative Adversarial Networks (GANs) and diffusion models introduce specific visual and spectral artifacts that can serve as forensic signals. GANs, for instance, often fail to replicate natural co-occurrence matrices \cite{zhang2019detecting}, and their outputs frequently exhibit high-frequency distortions due to upsampling layers \cite{durall2020watchupconvolutioncnnbased}.\\\\
In satellite imagery, these artifacts manifest in the following ways: (1) edge anomalies: abrupt transitions in roads or terrain, in contrast to the smooth gradients found in real data, and (2) spectral signatures: abnormal patterns in frequency domains (FFT/DCT), particularly with StyleGAN2-generated content. Multi-spectral approaches have been proposed to enhance forgery detection \cite{schmitt2018sen12}, but major benchmark datasets like DM-AER and FSI are limited to RGB channels, constraining the available forensic signals.\\\\
\textbf{2.5. Datasets and Evaluation Challenges}\\\\
Dataset quality and benchmarking protocols are central to the development of robust detection models. Yet, many existing studies suffer from limited reproducibility due to inconsistencies in metrics and evaluation strategies.\\\\
The DM-AER dataset \cite{deepmedia2022dmaer} contains over 1 million synthetic satellite images generated by StyleGAN2, offering significant diversity and complexity. The FSI dataset \cite{zhao2021deepfake} focuses on CycleGAN-generated forgeries and simulates military deception scenarios. These datasets serve as the foundation for our pipeline to training and evaluation.\\\\
To ensure generalization, we train on a combination of the DM‑AER and FSI datasets, allowing the model to learn from a diverse range of forgeries. This strategy draws inspiration from domain-specific practices in anomaly detection \cite{RAMADAN2024118718} and load forecasting using hybrid deep learning models \cite{10630814}.\\\\
\textbf{2.6. Explainability in Forensic Models}\\\\
For detection models to be useful in real-world contexts like defense or journalism, interpretability is essential. Model interpretability refers to the ability to understand and visualize how a model arrives at a decision, especially important when identifying subtle or distributed anomalies. CNN-based models often rely on Grad-CAM \cite{selvaraju2020gradcam}, which visualizes regions that most influenced the model’s output by computing gradients with respect to intermediate convolutional layers. In satellite forensics, this helps highlight manipulated areas like inconsistent shadows or duplicated buildings.\\\\
ViTs, however, require more sophisticated interpretability tools. Raw attention maps tend to be noisy or misaligned with true decision logic. To address this, \cite{chefer2021transformerinterpretabilityattentionvisualization} proposed a method that combines attention weights with gradient signals to produce high-resolution relevance maps. These maps offer more precise insights into which patches contribute most to forgery detection and how ViTs focus differently compared to CNNs.\\\\
\textbf{2.7. Research Gaps and Our Positioning}\\\\
Three major limitations persist in prior work:

1. lack of systematic comparison between CNNs and ViTs for satellite forgery detection,

2. over-reliance on qualitative visualizations with little quantitative artifact localization,

3. narrow evaluation pipelines, often using only one dataset or one model architecture.\\\\
Our study addresses these gaps by benchmarking both architectures across varied visual conditions and using cross-dataset evaluations to stress-test generalization. While we initially explored spectral artifact quantification, this component was ultimately excluded due to data constraints. We apply the interpretability method proposed by Chefer et al. \cite{chefer2021transformerinterpretabilityattentionvisualization} to Vision Transformers in the context of satellite deepfake detection, enabling a clearer understanding of model decisions. While we draw evaluation design inspiration from time-series transformer studies in solar forecasting \cite{RAMADAN2024118718}, our interpretability framework is based on vision-specific research.
\\\\
\section{Methodology} \label{sec:desc}
\noindent This section details the data sources, preprocessing steps, model architectures, training configurations, and evaluation procedures used to detect AI-generated satellite images using deep learning. We compare two model families, Convolutional Neural Networks (CNNs) and Vision Transformers (ViTs), trained on a large-scale dataset composed of real and synthetic satellite images. All experiments were implemented in PyTorch and logged using TensorBoard for monitoring performance.\\\\
\textbf{3.1 Dataset Composition}\\\\
We curated a combined dataset consisting of 138,064 satellite images drawn from two sources: the DM-AER dataset and the FSI dataset. The DM-AER dataset contributed the majority of the data, while 8,064 additional images were sourced from the FSI dataset to diversify the test and validation splits. All images were labeled as either real or fake, with fakes generated primarily via StyleGAN2 and CycleGAN pipelines. An 80-15-5 split was adopted for training, validation, and testing, respectively. Images were organized into three directories (train, val, and test) using the ImageFolder structure provided by the torchvision API.\\\\
\textbf{3.2 Preprocessing and Augmentation}\\\\
To enhance generalization and robustness, data augmentation was performed dynamically during training using the torchvision.transforms pipeline. Each image was randomly resized and cropped to 224×224 pixels, horizontally flipped, rotated by up to ±10°, and color-jittered to simulate real-world variability. This ensured that the model encountered slightly different versions of each image across epochs, promoting invariance to illumination and orientation.\\\\
For validation and testing, only resizing to 224×224 and normalization were applied. All images were normalized using the ImageNet mean and standard deviation:

\begin{equation*}
\mu = [0.485,\, 0.456,\, 0.406],
\end{equation*}
\begin{equation*}
\sigma = [0.229,\, 0.224,\, 0.225].
\end{equation*}

\noindent
This normalization was applied to align input distributions with the pretrained ImageNet backbones, ensuring stable convergence despite domain differences with satellite imagery. Together, these augmentations were implemented using torchvision.transforms and maintained consistent input formats across both CNN and ViT models.\\\\
\textbf{3.3 Model Architectures}\\\\
\textbf{3.3.1 Convolutional Neural Network (CNN)}\\\\
Our CNN-based model is based on the ResNet-50 architecture pretrained on ImageNet. To retain the benefit of learned hierarchical features while reducing overfitting, all convolutional layers were frozen during training. We experimented with both freezing and fine-tuning the backbone, and freezing consistently yielded better validation performance and more stable convergence, especially on the limited dataset size. The final classification layer was replaced with a custom head consisting of a dropout layer (p=0.4) followed by a fully connected layer with two output nodes corresponding to binary classification. This structure helped mitigate overfitting while enhancing generalization to unseen fakes. The full model was trained on a GPU using mixed-precision support when available\\\\
\textbf{3.3.2 Vision Transformer (ViT)}\\\\
The ViT model used in our experiments is the ViT-B/16 architecture implemented via the timm library \cite{rw2019timm}. The pretrained backbone was frozen to preserve transformer-based global pattern representations, and only the classification head was fine-tuned on our dataset. This decision was made after empirical testing; fine-tuning the entire transformer led to overfitting and less stable training, whereas freezing the backbone consistently resulted in higher validation accuracy and F1-scores. Like the CNN, the head consisted of a fully connected layer adapted for binary classification. The ViT approach is particularly suitable for identifying long-range dependencies and spatial inconsistencies, such as duplicated terrain and unnatural road layouts.\\\\
\textbf{3.4 Training Configuration}\\\\
Both models were trained for 20 epochs with a batch size of 32 using the Adam optimizer. The learning rate was set to 1e-4 with a weight decay of 1e-5 to discourage overfitting. To improve convergence and prevent stagnation, we utilized a ReduceLROnPlateau scheduler with a patience of 3 and a decay factor of 0.5. The loss function used was cross-entropy loss. Training was conducted using the PyTorch framework, and model checkpoints were saved whenever a new best validation accuracy was recorded.\\\\
Throughout training, metrics such as loss, accuracy, precision, recall, and F1-score (macro-averaged) were computed per epoch and logged using TensorBoard. This enabled real-time monitoring of model behavior and convergence dynamics.\\\\
\textbf{3.5 Evaluation Strategy}\\\\
After training, each model was evaluated on the held-out test set to measure generalization performance. The test set included samples from both DM-AER and FSI datasets, ensuring cross-domain variability. Key evaluation metrics included overall classification accuracy, macro-averaged precision, recall, and F1-score.\\\\
To assess model robustness, we performed qualitative error analysis focusing on commonly observed deepfake artifacts such as repeated textures, terrain inconsistencies, and shadow mismatches. The CNN was expected to focus on local irregularities, whereas the ViT was hypothesized to identify broader contextual anomalies, consistent with findings from interpretability studies.
\\\\
TensorBoard visualizations were also used to analyze learning trends, model confidence, and class-wise prediction behavior, facilitating transparency and reproducibility.\\\\
\section{Evaluation and Results}  \label{sec:fdesc}
\noindent The trained CNN and ViT models were evaluated on the held-out test set, which included samples from both DM-AER and FSI datasets. As illustrated in Figures 2–4, the Vision Transformer (ViT) consistently outperformed the ResNet-50 CNN across all validation metrics.\\\\
Figure 2 shows that ViT achieved higher macro-averaged F1-scores, precision, and recall throughout training, maintaining more stable values across epochs.\\\\
\begin{figure}[H]
    \centering
    \includegraphics[width=0.8\linewidth]{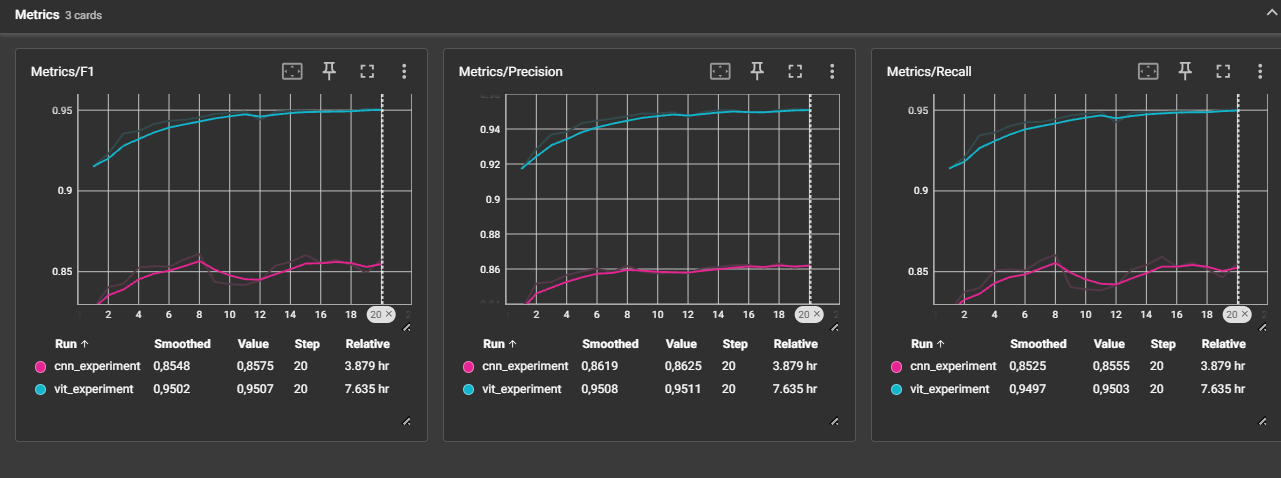}
    \caption{F1-score, precision, and recall comparison  over epochs between CNN and ViT.}
    \label{fig:metricslikef1andetc}
\end{figure}
\noindent
Figure 3 presents the training and validation loss trajectories, where ViT converged faster and reached a lower final validation loss ($\approx 0.134$) compared to CNN ($\approx 0.336$).\\\\
\begin{figure}[H]
    \centering
    \includegraphics[width=0.8\linewidth]{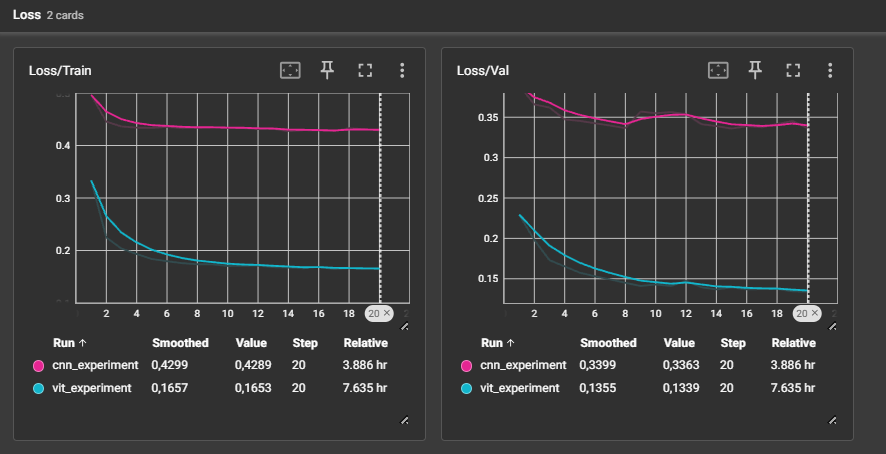}
    \caption{Training and validation loss curves for ViT and CNN on 20 epochs.}
    \label{fig:metricslikef1andetc}
\end{figure}
\noindent
Figure 4 displays the accuracy curves, highlighting ViT’s superior generalization during validation ($\approx 95\%$ vs. $86\%$ for CNN).\\\\
\begin{figure}[H]
    \centering
    \includegraphics[width=0.8\linewidth]{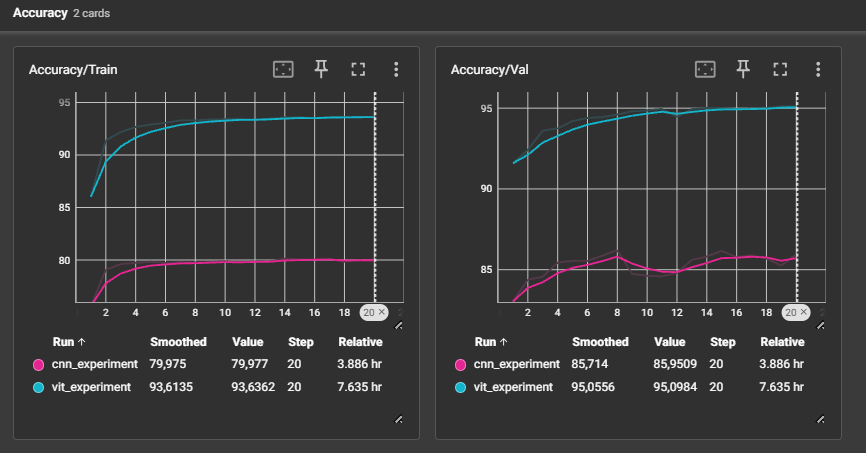}
    \caption{Accuracy over epochs for CNN and ViT models on the training and validation sets.}
    \label{fig:metricslikef1andetc}
\end{figure}
\noindent
These results collectively demonstrate ViT’s advantage in modeling long-range dependencies and global structure. Quantitatively, ViT achieved higher validation accuracy ($95.1\%$ vs. $86.0\%$), F1-score (0.951 vs. 0.857), precision (0.951 vs. 0.863), and recall (0.950 vs. 0.856).\\\\
Beyond validation performance, both models were further evaluated on the held-out $5\%$ test split to assess generalization. The ViT achieved a final test accuracy of $95.11\%$, while the CNN reached $87.02\%$, confirming that the transformer’s improvements extend beyond the training and validation phases.\\\\
For completeness, we also trained a DenseNet-121 model under the same configuration as ResNet-50 to test whether ViT’s advantage extends beyond a single CNN design. DenseNet reached only $83\%$ validation accuracy and an F1-score of 0.835, performing below both ResNet-50 and ViT. This further confirms the broader superiority of transformer-based architectures for detecting AI-generated satellite imagery.\\\\
To further interpret model performance, we examined confusion matrices and representative misclassifications. Both models achieved high true-positive and true-negative rates overall. The CNN’s confusion matrix displayed a modest number of false positives and negatives, while the ViT’s matrix was nearly diagonal, indicating very few errors. The CNN occasionally misclassified visually uniform regions such as deserts, cloudy ocean areas, or heavily noisy tiles. Although these images were authentic, their lack of distinctive structure made them visually similar to GAN-generated artifacts. The ViT, by contrast, handled these cases more reliably due to its global attention mechanism and stronger contextual reasoning.\\\\
Insights from the confusion matrices also informed model tuning. We used them to identify artifact types associated with false classifications, leading to targeted augmentation adjustments and threshold refinements. These modifications reduced error propagation in later training stages. Across all evaluations, the ViT maintained its edge in generalization, showing greater robustness to edge cases and distribution shifts. These findings validate our hypothesis that self-attention mechanisms are inherently more effective at capturing the spatial and semantic irregularities that characterize deepfake satellite imagery.\\\\
\section{Analysis} \label{sec:conclusion}
\noindent To interpret the decision-making processes of the trained models, we employed post-hoc explainability techniques tailored to the architecture of each network family. These methods allowed us to visualize which regions of input satellite images were most influential in driving the models’ predictions, helping validate model reliability and reveal potential blind spots.\\\\
\textbf{5.1 CNN Analysis: Grad-CAM Saliency Maps}\\\\
To interpret the CNN model's decision-making process, we applied Gradient-weighted Class Activation Mapping (Grad-CAM) to the final convolutional layer (layer4[-1]) of the pretrained ResNet-50. Using a custom wrapper around this layer, we generated heatmaps for selected real and fake test samples, highlighting which image regions contributed most to the classification output. The Grad-CAM implementation was performed in PyTorch using OpenCV and Matplotlib for overlay visualization.\\\\
Images were first transformed using the same pipeline as during training (resize to 224×224, normalization), then passed through the model in evaluation mode. Heatmaps were overlaid on the original RGB image to visualize activation intensity. Figure 5 shows an example output: The CNN strongly responded to boundary artifacts, such as inconsistent lake edges, unnatural blending with nearby urban areas, and irregular vegetation textures, features commonly associated with GAN-based image synthesis. \\\\

\begin{figure}[H]
    \centering
    \includegraphics[width=0.8\linewidth]{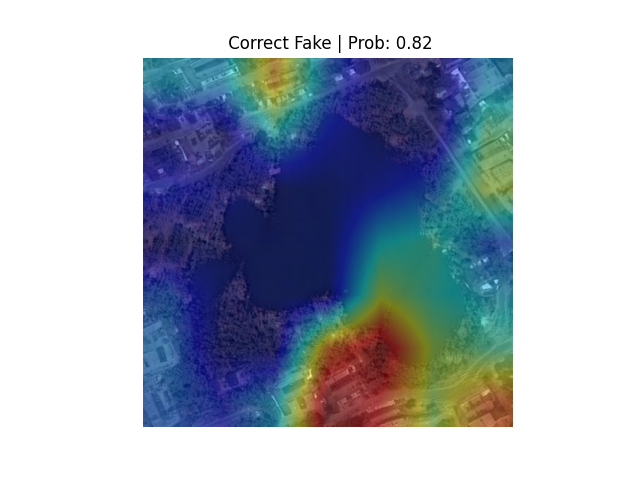}
    \caption{Grad-CAM visualization of a fake satellite image classified by the CNN model.}
    \label{fig:example_gradcam}
\end{figure}
\noindent
However, consistent with expectations, the Grad-CAM maps showed limited global awareness. In real images with sparse features (e.g., deserts, oceans), the model sometimes made incorrect predictions, likely due to insufficient local cues and lack of contextual understanding.\\\\
\textbf{5.2 ViT Analysis: Attention Attribution via Chefer et al. (2021)}\\\\
To visualize attention attribution in the Vision Transformer, we employed the Chefer et al. (2021) rollout-based explainability method. This technique propagates relevance scores through the attention layers of the ViT model (vit\_base\_patch16\_224) and generates high-resolution attention masks that emphasize globally important regions of the image.\\\\
We implemented this using a VitAttentionRollout module, with discard\_ratio=0.3 to preserve salient attention paths and head\_fusion='mean' to integrate multi-head information. Each image was resized to 224×224 pixels, normalized with the ImageNet mean and standard deviation for consistency with the CNN preprocessing, and passed through the ViT to generate predictions paired with their corresponding attention heatmaps from the rollout process. The attention mask was resized to match the original image and blended using alpha compositing.\\\\
Figure 6 illustrates a ViT attention attribution map for a synthetic industrial site classified as fake with $99.88\%$ confidence. The attention heatmap shows that the model focuses primarily on the circular storage tanks and grid-like network of pipelines distributed throughout the facility. These areas, highlighted by brighter regions in the overlay, reflect the ViT’s sensitivity to the overly uniform spacing and repetitive geometry typical of GAN-generated industrial patterns. Unlike CNNs that emphasize small texture patches, the ViT captures global spatial relationships, recognizing how identical shapes and regular alignments extend across the entire layout. Even under muted color and lighting conditions, the model maintains attention on these structural cues, explaining its robustness in recognizing synthetic imagery.\\\\
\begin{figure}[H]
    \centering
    \includegraphics[width=0.5\linewidth]{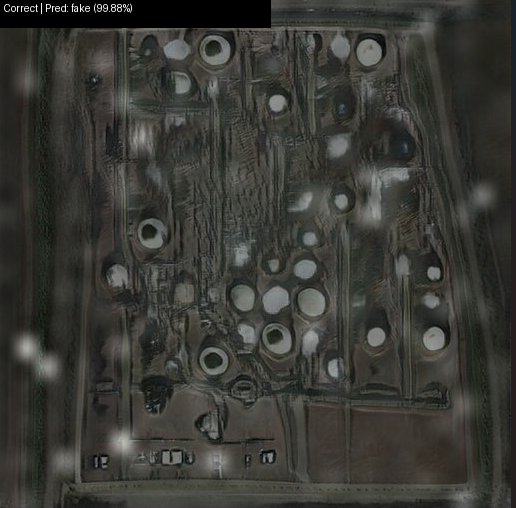}
    \caption{Vision Transformer (ViT) attention attribution on a synthetic industrial site. Bright regions correspond to the model’s focus on globally repetitive layouts and circular structures indicative of synthetic generation.}
    \label{fig:example_chefervit}
\end{figure}
\noindent
\textbf{5.3 Visual Comparison of CNN and ViT Explanations}\\\\
To compare how each architecture interprets visual information, we applied Grad-CAM to the CNN (ResNet-50) and the Chefer et al. (2021) attention-rollout method to the ViT (ViT-B/16). Both techniques produce heatmaps that highlight the image regions most influential to the model’s classification.\\\\
We selected 20 samples for the CNN (10 correctly classified and 10 misclassified) and 8 samples for the ViT (4 correctly classified, 4 misclassified). Each image was overlaid with its corresponding heatmap to visualize model focus.\\\\
\textbf{A. Correct Predictions}\\\\
When both models correctly classified the input (real or fake), their explanations reveal complementary detection behaviors.

\begin{itemize}
    \item Grad-CAM (CNN): Focused on localized regions with repetitive textures and terrain distortions, such as building edges or patterned vegetation.
    \item ViT Attention: Distributed more broadly, capturing structured layouts and overall spatial organization across the scene.
\end{itemize}
Interpretation: The CNN relies on fine-grained texture cues, whereas the ViT integrates global context such as road curvature, vegetation alignment, and structural symmetry. This distinction is evident in scenes like urban blocks and parking areas (Figure 7a for ViT; Figure 8a for CNN), where the ViT spreads attention across the entire structure, while the CNN concentrates on localized details.\\\\
\begin{figure}[H]
    \centering
    \includegraphics[width=0.8\linewidth]{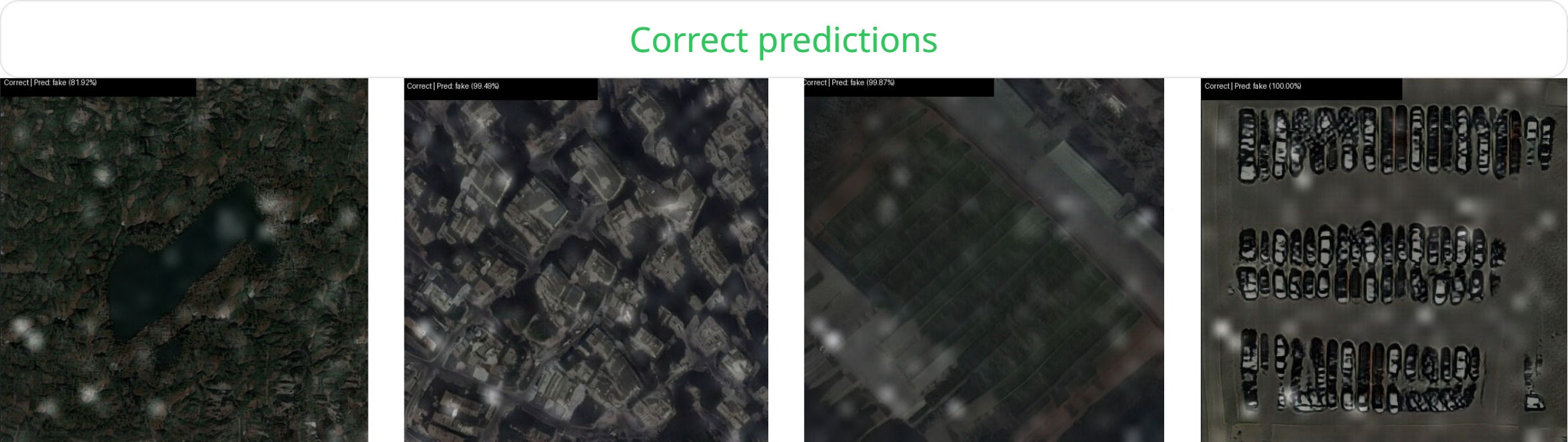}
    \caption{Vision Transformer (ViT) attention rollout visualizations.
    (a) Correctly classified samples. (b) Misclassified samples (shown later in Section 5.3 B).}
    \label{fig:correct_vits}
\end{figure}

\begin{figure}[H]
    \centering
    \includegraphics[width=0.8\linewidth]{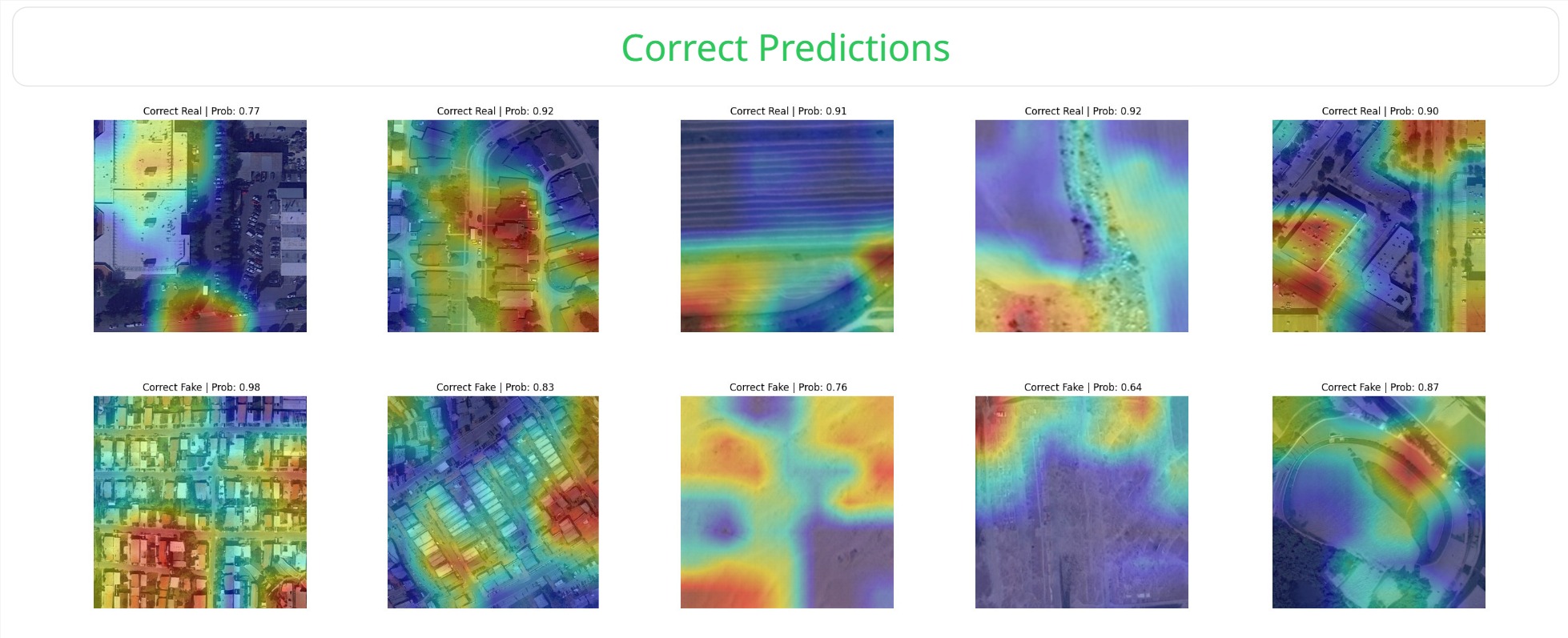}
    \caption{CNN (ResNet-50) Grad-CAM saliency maps.
     (a) Correctly classified samples. (b) Misclassified samples (shown later in Section 5.3 B).}
    \label{fig:correct_gradcam}
\end{figure}
\noindent
\textbf{B. Incorrect Predictions}\\\\
When one or both models misclassified the input, their visualization patterns revealed distinct weaknesses.
\begin{itemize}
    \item CNN Errors: The CNN frequently misclassifies low-texture real images as fake or fails when GAN-generated samples are overly smooth and lack strong edge artifacts. Grad-CAM maps become diffused or highlight irrelevant regions.
    \item ViT Errors: ViT mistakes are rarer but occur in visually ambiguous scenes containing mixed semantics or extreme class imbalance. Even in such cases, attention remains broadly aligned with relevant structural regions.
\end{itemize}
Interpretation: The CNN’s dependence on localized textures makes it vulnerable to smooth or blended surfaces, whereas the ViT’s global attention is more semantically grounded but can still falter when foreground objects (e.g., buildings or forests) appear realistic while background inconsistencies exist. (Figure 7b for ViT; Figure 8b for CNN) illustrate these failure cases\\\\
\begin{figure}[H]
    \centering
    \addtocounter{figure}{-2}
    \renewcommand{\thefigure}{\arabic{figure}(b)}
    \includegraphics[width=0.8\linewidth]{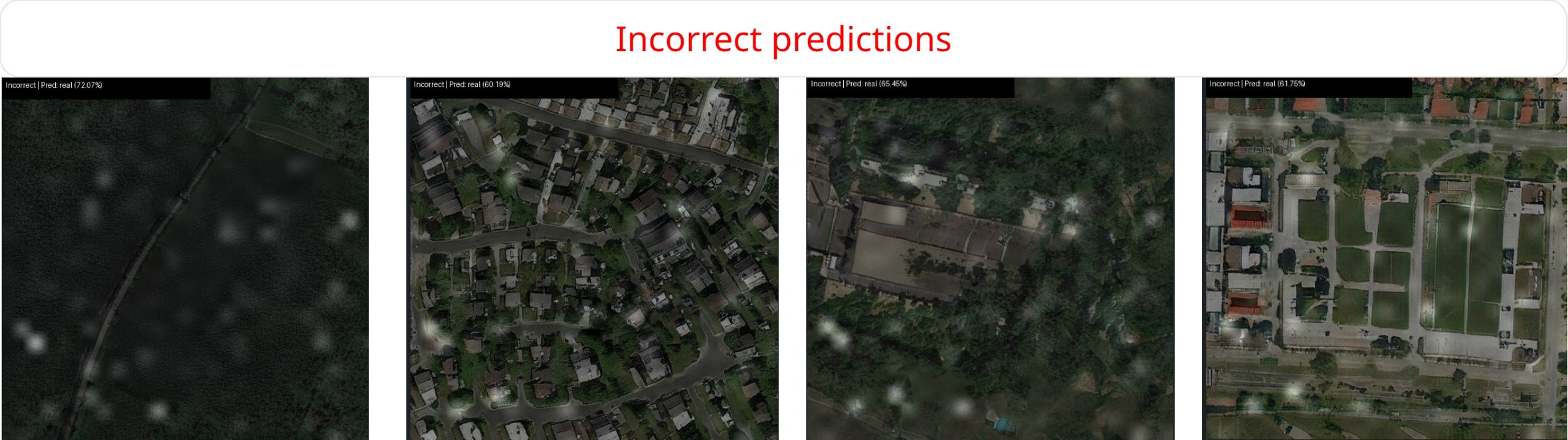}
    \caption{ViT attention rollout for misclassified samples.}
    \label{fig:7b}
\end{figure}

\begin{figure}[H]
    \centering
    \renewcommand{\thefigure}{\arabic{figure}(b)}
    \includegraphics[width=0.8\linewidth]{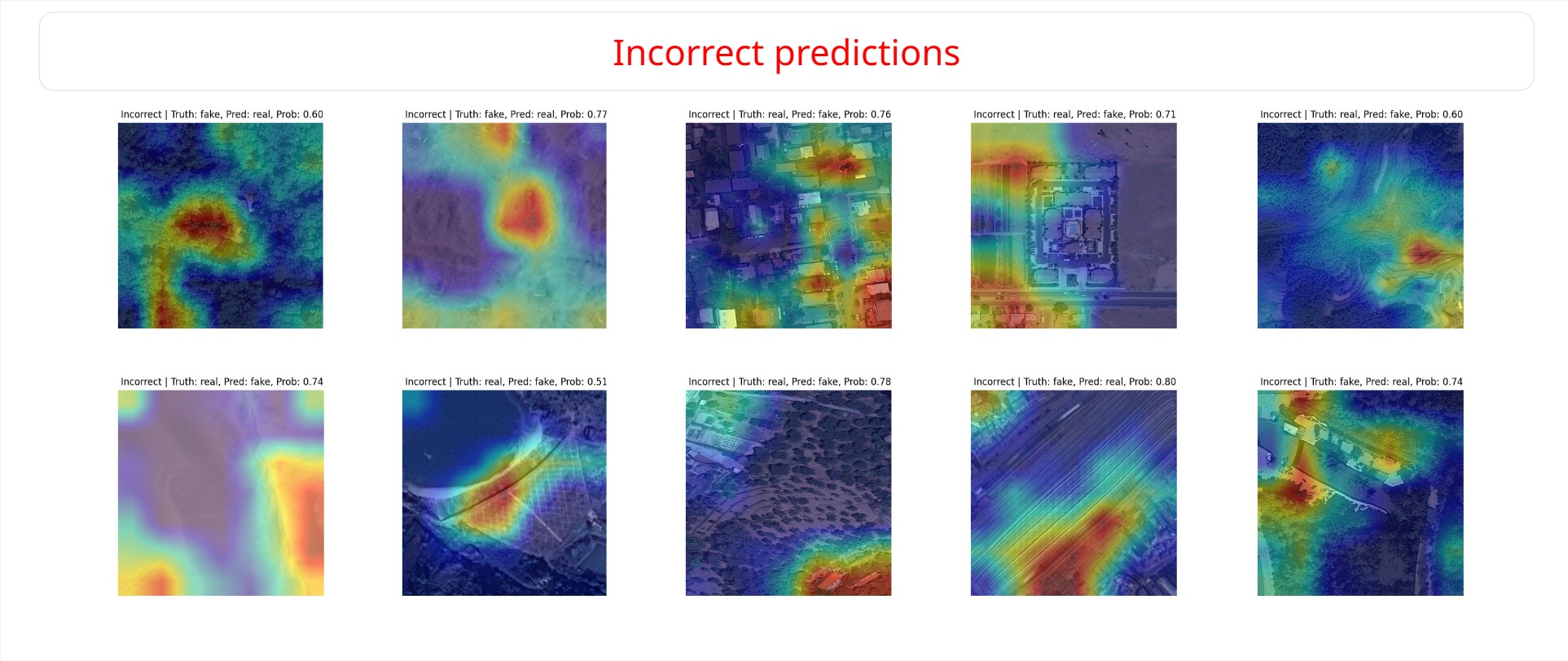}
    \caption{CNN Grad-CAM saliency maps for misclassified samples.}
    \label{fig:incorrect_gradcam}
\end{figure}
\noindent
\textbf{Summary}\\\\
These visual comparisons reinforce the quantitative findings:
CNNs excel at detecting texture-based artifacts but are constrained by their local receptive fields, while ViTs, through global self-attention mechanisms, demonstrate stronger robustness to spatial manipulation and semantic inconsistencies, making them better suited for satellite deepfake detection.
\\\\
\section{Conclusion} 
\noindent As generative models advance, preserving the authenticity of satellite imagery is vital to maintain trust in remote-sensing data and derived analyses. The rapid progress of these models introduces new risks of large-scale misinformation and forged geospatial content, emphasizing the need for reliable detection methods.\\\\
This study presents a systematic evaluation of deepfake detection in satellite imagery, comparing the effectiveness of Convolutional Neural Networks (CNNs) and Vision Transformers (ViTs) on a mixed-domain dataset derived from DM-AER and FSI. The primary aim was to determine which architecture is better suited for identifying AI-generated satellite forgeries and to explore how interpretability tools can enhance model trust and transparency.\\\\
The Vision Transformer (ViT-B/16) consistently outperformed the CNN (ResNet-50) across all evaluation metrics. ViT achieved a test accuracy of $95.11\%$ and a macro-averaged F1-score of 0.951, demonstrating greater robustness under distributional variation. In contrast, the CNN peaked at $87.02\%$ test accuracy and 0.857 F1-score, showing vulnerability to low-texture or ambiguous regions.\\\\
ViT's superiority stems from its ability to model long-range dependencies and capture spatial coherence, a critical factor in satellite imagery where anomalies often span large contextual regions. While CNNs excel at detecting localized artifacts such as repeated textures or sharp edge transitions, they lack the global perspective necessary to identify macro-level inconsistencies like terrain duplications or mismatched infrastructure. ViTs, equipped with self-attention mechanisms, demonstrated enhanced semantic reasoning and reduced susceptibility to context-poor misclassifications.\\\\
The most prominent visual artifacts were repetitive patterns, unnatural terrain transitions, abrupt road terminations, and shadow inconsistencies. Spectrally, GAN-generated images introduced high-frequency distortions, while diffusion-based forgeries often exhibited semantic dissonance, such as disconnected structures or illogical object placements. These inconsistencies were especially prevalent in StyleGAN2-generated outputs and were effectively leveraged by both model architectures to distinguish real from synthetic imagery.\\\\
Interpretability played a crucial role in validating model behavior and enhancing transparency. Grad-CAM for CNNs localized fine-grained irregularities, such as texture repetition or boundary artifacts, but lacked a holistic view. In contrast, the Chefer et al. \cite{chefer2021transformerinterpretabilityattentionvisualization} attribution method applied to ViTs yielded more informative, high-resolution relevance maps, capturing global structural cues and contextual anomalies with greater precision. These visualizations not only confirmed the ViT’s attention to meaningful spatial patterns but also offered forensic analysts a valuable window into the model’s decision-making logic.\\\\
Despite strong results using RGB data, current methods remain constrained by dataset limitations. In future work, we plan to explore synthetic detection in Synthetic Aperture Radar (SAR) and multispectral satellite imagery, which can offer deeper physical and spectral insights unavailable in RGB alone. Developing large-scale labeled datasets in these modalities will be essential. Additionally, integrating frequency-domain techniques such as Fast Fourier Transform (FFT) and Discrete Cosine Transform (DCT) may further enhance the model's sensitivity to periodic or compression-based artifacts often introduced by generative models. These extensions could improve both detection accuracy and generalizability in operational settings where image authenticity is critical.\\\\

\begin{singlespace}
\bibliographystyle{style}

\bibliography{biblio}
\end{singlespace}

\end{document}